\documentclass[11pt,a4paper]{article}
\usepackage[hyperref]{acl2020}
\usepackage{times}
\usepackage{latexsym}
\usepackage{graphicx}
\usepackage{makecell}
\usepackage{multirow}
\usepackage{url}
\usepackage{subcaption}
\usepackage{booktabs}
\usepackage{enumitem}
\usepackage{algorithm}
\usepackage[noend]{algpseudocode}
\usepackage[colorinlistoftodos,prependcaption,textsize=tiny]{todonotes}
\usepackage{float}
\usepackage{color, colortbl}
\usepackage{mdframed}
\usepackage{xcolor, soul}  
\usepackage{float}
\usepackage{amsmath}
\usepackage{setspace}

\usepackage{eqparbox}

\algrenewcommand\algorithmicindent{.9em} 

\usepackage{listings}
\aclfinalcopy 

\title{TALM: Tool Augmented Language Models}

\author{
Aaron Parisi\qquad Yao Zhao\qquad Noah Fiedel\\
\\
  {\tt \{aarontp$, $yaozhaoyz$, $nfiedel\}}
  \\
  
  {\tt @google.com}}
\date{}

\begin{document}
\maketitle

\begin{abstract}

Transformer based language models (LMs) demonstrate increasing performance with scale across a wide variety of tasks.  Scale alone however cannot enable models to solve tasks that require access to ephemeral, changing, or private data that was unavailable at training time. Many useful tasks may also benefit from LMs being able to access APIs that read or modify state. In this work, we present Tool Augmented Language Models (TALM), combining a text-only approach to augment language models with non-differentiable tools, and an iterative ``self-play" technique to bootstrap performance starting from few tool demonstrations. TALM exhibits strong performance on both a knowledge-heavy QA task and a reasoning oriented math task with simple tools. At a given model scale, TALM significantly outperforms  non-augmented LMs. We further demonstrate that TALM successfully performs out-of-distribution inferences on both QA and math tasks, where non-augmented LMs fail. Our results suggest that Tool Augmented Language Models are a promising direction to enrich LMs' capabilities, with less dependence on scale.


\end{abstract}

\section{Introduction}

Language models using the Transformer architecture \citep{vaswani2017attention} demonstrate increasing performance at larger scales, e.g. T5 \citep{raffel2019t5}, GPT-3 \citep{brown2020gpt3}, and PaLM \citep{chowdhery2022palm}. Scale related performance gains are observed on a variety of benchmarks, e.g. SuperGLUE \citep{wang2019superglue} and BIG-bench \citep{bigbench}.

Scaling up has practical downsides. Large scale models are unwieldy to store, transfer, and deploy. Their costs to train or perform inference can be prohibitively high for many researchers and organizations.

\begin{figure}[tbh]
\centering
\includegraphics[width=1.0\linewidth, trim=80 0 80 0]{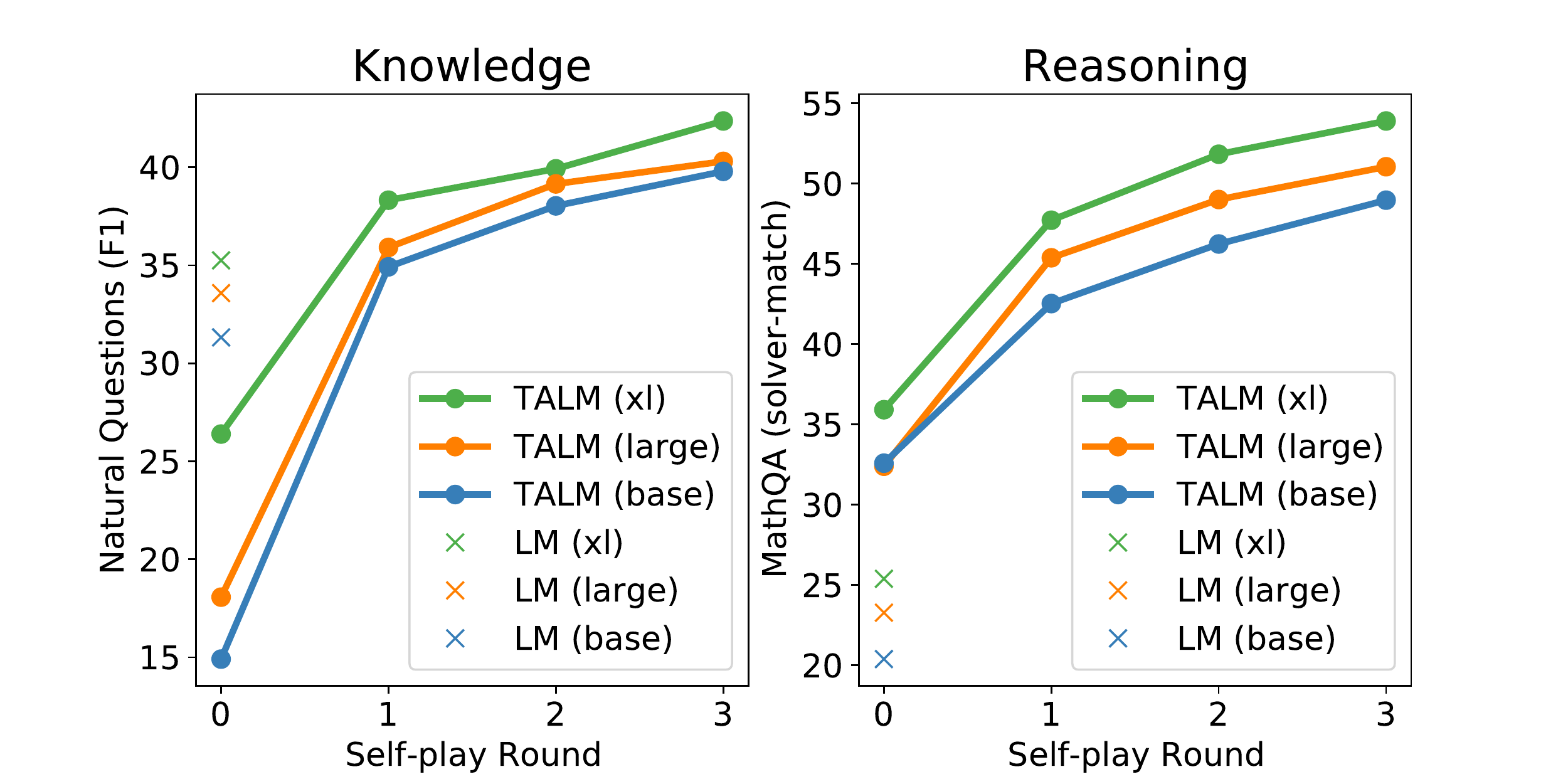}
\caption{Baseline LM and TALM performance on two tasks, with increasing rounds of self-play.}

\label{fig:self-play-round}
\vspace{-1em}
\end{figure}

Larger models memorize more world knowledge \citep{adarob2020cbqa}. While good for many benchmark tasks, relying on memorization alone poses several problems. First, models sometimes generate incorrect outputs that are problematic for some applications. Second, world knowledge is constantly changing. The knowledge from yesterday’s training data might be invalid today. Third, large models can memorize parts of their training data with undesirable consequences \citep{carlini2022memorization}.

Retrieval based approaches to enhancing LMs can lower the dependence on scale. REALM \citep{guu2020realm} learns retrieval via backpropagation from a fixed corpus. RETRO \citep{borgeaud2021improving} adds an "internet scale" retrieval mechanism. RAG \citep{lewis2020retrieval} uses a dense vector index of Wikipedia, and retrieves either once per token or once per query. Other works demonstrated that LMs can be enhanced on math reasoning with access to a calculator \cite{andor2019giving}.

Looking towards the future utility of language models, it is clear that scale and retrieval cannot solve all useful problems. Many knowledge tasks and desirable applications require access to read live or private data (e.g. weather or a person's calendar), or to invoke APIs that modify state. Recent works such as Say Can \citep{ahn2022saycan} connect languages models to an environment, though with the model as a recipient of queries. In contrast, TALM's approach enables models to invoke arbitrary tools with model-generated output, and to attend to tool output to generate task outputs.


In summary, our contributions are:
\vspace{-.5em}
\begin{itemize}
  \setlength\itemsep{0em}
  \item Demonstrating that language models can be augmented with tools via a text-to-text API.
  
  \item Demonstrating an iterative self-play technique to bootstrap tool-augmented datasets and subsequent tool-augmented model performance, from few labeled examples.

\end{itemize}

\section{Methods}

We use pretrained T5 models \citep{raffel2019t5, roberts2022t5x} for finetuning, inference and evaluation. To measure the effects of model scaling, we use the base, large, and XL sizes.

\subsection{Tool Augmented Language Models}

\begin{figure}[tbh]
\centering
\includegraphics[width=1.0\linewidth, trim=0 10 0 10]{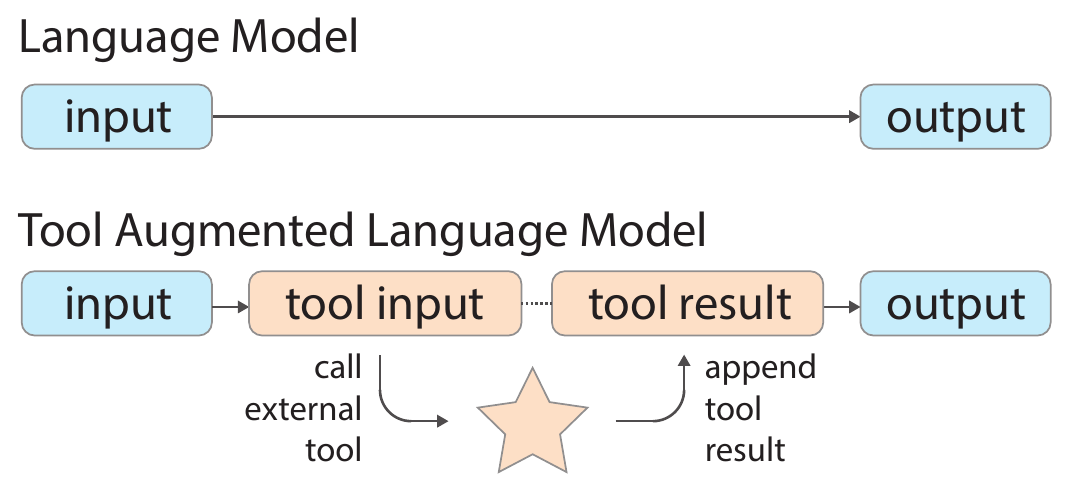}
\caption{LM and Tool Augmented LMs.}

\label{fig:talm-comparison}
\end{figure}

We use a Text-to-Text tool interface given its broad applicability and simplicity, as shown in Fig.~\ref{fig:tasks_example3}.
TALM first generates a tool input conditioned on the task input text and invokes a tool's API by generating a delimiter, such as "\textit{\textbar result}". Whenever this delimiter is detected, the tool API is called and its result appended to the text sequence. TALM then continues to generate the final task output.

\begin{figure}[h!]
\centering

\begin{mdframed}
\small

\textbf{An abstract task:}

\textcolor{teal}{task input text} 
\textcolor{orange}{\emph{\textbf{\textbar tool-call}} tool input text} 
\textcolor{blue}{\emph{\textbf{\textbar result}} tool output text} \textcolor{violet}{\emph{\textbf{\textbar output}} task output text}

\end{mdframed}



\begin{mdframed}
\small

\textbf{A weather task:}

\textcolor{teal}{how hot will it get in NYC today?} 
\textcolor{orange}{\emph{\textbf{\textbar weather}} lookup region=NYC} 
\textcolor{blue}{\emph{\textbf{\textbar result}} precipitation chance: 10, high temp: 20c, low-temp: 12c}
\textcolor{violet}{\emph{\textbf{\textbar output}} today's high will be 20C}

\end{mdframed}
\vspace{-10pt}
\caption{TALM text-to-text interface example.}
\label{fig:tasks_example3}
\end{figure}

TALM learns two subtasks at the same time: calling a tool and generating an answer based on tool results. TALM is architecturally agnostic and can be implemented as Seq2Seq, left-to-right LM or prefix LM. 
We chose the Seq2Seq family for its high finetuning performance at modest scale \cite{raffel2019t5}.

\subsection{Iterative self-play}
\label{sec:self-play}

When introducing new tools to solve existing tasks,
there are often a limited number of demonstrations of tool interactions.
However, there is typically plenty of supervised task data consisting of input and target pairs, and automated metrics for evaluating the correctness of a generated output. Inspired by Decision Transformer \citep{chen2021decision}, we use a self-play approach to iteratively bootstrap examples of tool-use with progressively higher quality. In this work, we refer to a model interacting with a tool API as self-play rather than adversarial play among models.

\begin{algorithm}
  \small
  \begin{spacing}{1.3}
  \begin{algorithmic}[1]

    \State $T=\{x_i, y_i\}_T$ \Comment{task set} 
    \State $D=\{x_j, t_j, r_j, y_j\}_D$\Comment{tool-use set}
    \State $P_\theta \gets pretrained~LM$

    \For{$t \in [0, 1, ..., R]$} \Comment{self-play rounds} 
        \State \Comment{finetune LM} 
        \State $\theta \gets \underset{\theta}{\operatorname{argmax}}\prod_{D}  P_{\theta}(y_j|x_j, t_j, r_j)P_{\theta}(t_j|x_j)$
      \For{$x_i, y_i \in T$} \Comment{iterate task set} 
        \For{$n \in [0, 1, ..., N]$}
            \State $t_n \gets P_{\theta}(t|x_i)$ \Comment{sample tool query} 
            \State $r_n \gets Tool(t_n)$ \Comment{call tool API} 
            \State $y_n \gets P_{\theta}(y|x_i,t_n,r_n)$ \Comment{get task output} 
                \If{$|y_n-y_i|<th$ }  \Comment{filter wrong output} 
                    \State $D \gets D \cup \{x_i, t_n, r_n, y_n\}_1$
                    \State \Comment{update tool-use set} 
                \EndIf
        \EndFor
      \EndFor
    \EndFor
  \end{algorithmic}
  \end{spacing}
\label{algorithm-iterative}
\caption{Iterative Self-Play Algorithm. \newline \small{$x$: task input, $y$: task output, $t$: tool input, $r$: tool output}}
\end{algorithm}

The iterative self-play pipeline starts with a small tool-use bootstrapping set $\{x_j, t_j, r_j, y_j\}_D$. In each round of self-play, the TALM is fine-tuned on the tool-use set $D$. Next, for every example in the task set $T$, the TALM samples tool inputs, calls a tool API, and samples task outputs based on the tool results. If the TALM generated task output matches the target within some threshold $th$, the tool-use sequence led to the result is added to the tool-use set $D$ for the next round of self-play.

To explore diverse tool API invocations and answers during self-play, the TALM decodes using random sampling with temperature=1.0, and top-k=40. To grow the dataset during self-play, the TALM generates up to $N$=600 tool-use sequences per example. At evaluation time, the model uses beam decoding with 4 beams to generate a single output.

We note that this iterative self-play pipeline represents a special case of a policy-gradient RL algorithm, where the LM is the policy network and is trained by policy gradient with a binary reward signal. Iterative self-play is related to expert iteration \cite{anthony2017expertiteration}, which has been demonstrated to work well in tasks with extremely weak supervision \cite{christiano2018amplification}. While our tasks are currently single-hop, this formulation can be extended further into RL: modelling multihop tool-use tasks as markov decision processes (MDPs), or integrating algorithms like Decision Transformer \cite{chen2021decision}.

\section{Results}
\label{results-section}

We evaluate TALM on two domains. The first is the knowledge-oriented Natural Questions (NQ) \citep{kwiatkowski2019natural}, a diverse QA task. The second is MathQA \citep{amini2019mathqa}, selected to measure general reasoning capability rather than knowledge.


\subsection{Natural Questions}
Natural Questions (NQ) is a large ($\approx$ 300k training examples) QA dataset collected from real user queries. NQ contains both long and short answer tasks. We selected the short answer task as it is both more challenging as measured with lower baseline performance, and closer to practical use-cases such as assistants. In addition to a question and short-answer pair, examples in the NQ dataset include an "oracle" context (span) of a Wikipedia document containing the answer. We remove boolean questions to avoid inflated performance due to random-chance guesses. We compare TALM against closed-book LM benchmarks. 

For TALM experiments, we do not feed the oracle contexts directly to the model, instead using them to populate an index that TALM can access as a retrieval tool. The retrieval system is implemented using a BM25-based index over the union of all NQ oracle contexts.

\begin{figure}[h!]
\centering

\begin{mdframed}
\small

\textbf{Question:}
\textcolor{violet}{when are hops added in brewing process?} 
\vspace{3pt}

\textbf{Short Answer:}
\textcolor{blue}{The boiling process.}


\end{mdframed}

\begin{mdframed}
\small

\textcolor{violet}{$|$question when are hops added in brewing process?} 
\textcolor{teal}{$|$search brewing process}
\textcolor{teal}{$|$result The boiling process is where chemical reactions take place...including}
\textcolor{blue}{$|$output The boiling process.}


\end{mdframed}

\vspace{-5pt}

\caption{Example from Natural Questions, as a standard NQ task and the corresponding tool-augmented sequence.}
\label{fig:nq_example}
\end{figure}

\begin{figure}[tbh]
\centering
\includegraphics[trim=0 0 50 50, width=0.9\linewidth]{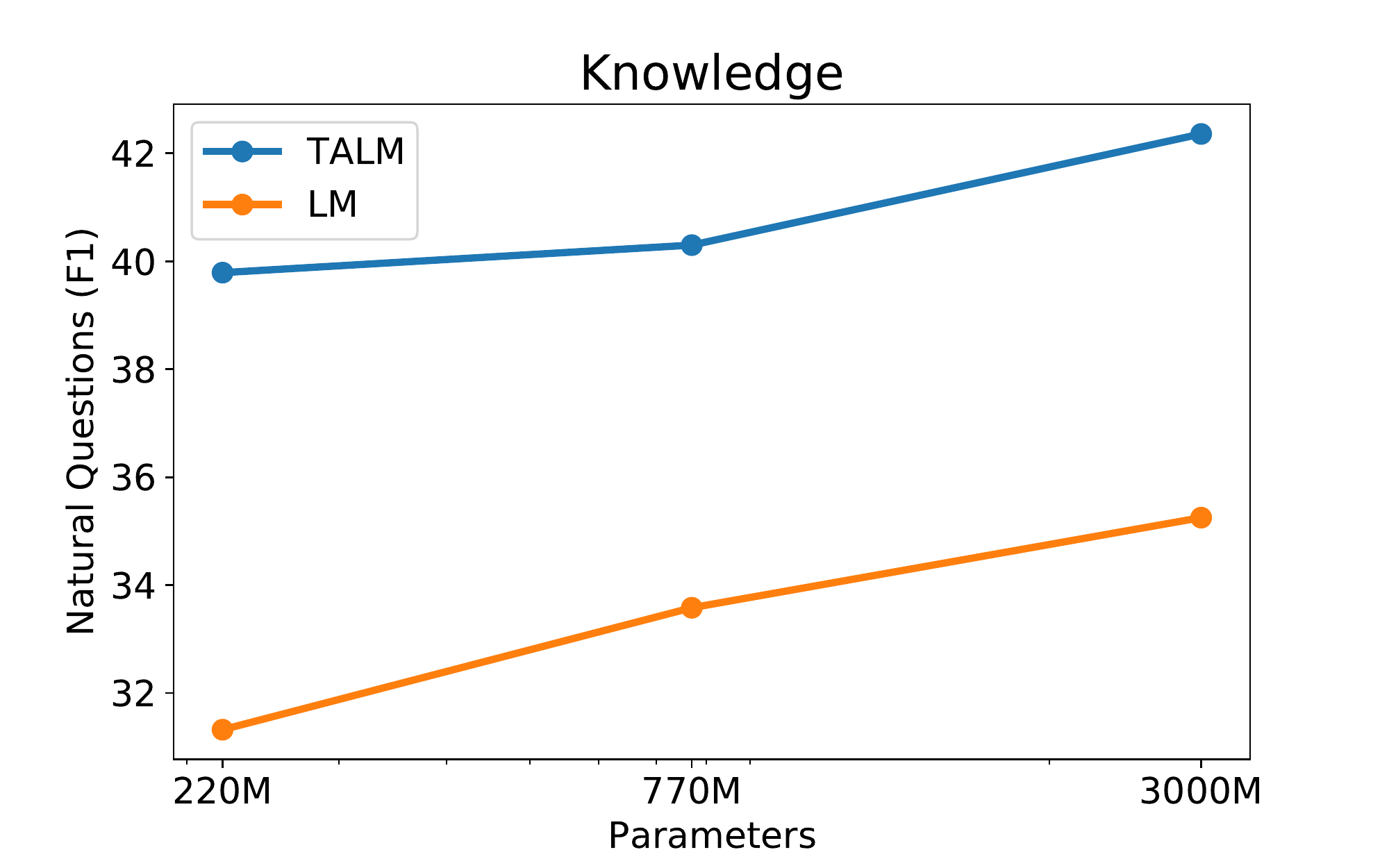}
\caption{Performance of TALM compared with TALM of different model sizes on Natural Questions. The TALM is bootstrapped from 150 tool demonstrations and undergoes two rounds of self-plays. We hypothesize that the noise in the performance-scale curves is due to finetuning in a low-data regime.}
\label{fig:nq_talm}
\end{figure}


In Fig.~\ref{fig:nq_talm}, even the $220M$ base TALM outperforms $3B$ XL LM.
There is also a smaller performance gap between base and XL sized TALMs than between TALM and LM, suggesting that smaller models benefit more from retrieval tools for knowledge intensive tasks.


\subsection{MathQA}
MathQA \citep{amini2019mathqa} is a large scale dataset of math word problems ($\approx 30k$ training examples). Each example includes the word problem, a formula generated by crowd-source workers to calculate the answer, and the correct text-form answer among multiple choices. 

\begin{figure}[h!]
\centering

\begin{mdframed}
\small

\textbf{Question:}
\textcolor{violet}{If Lily's test scores are 85 , 88 and 95 out of 100 in 3 different subjects , what will be her average score?} 
\vspace{3pt}

\textbf{Formula:}
\textcolor{orange}{Divide(Add(85, Add(88, 95)), 3)} 
\vspace{3pt}

\textbf{Answer:}
\textcolor{blue}{89.33}


\end{mdframed}

\begin{mdframed}
\small

\textcolor{violet}{$|$question If Lily's test scores are 85 , 88 and 95 out of 100 in 3 different subjects , what will be her average score?} 
\textcolor{teal}{$|$formula Divide(Add(85, Add(88, 95)), 3)}
\textcolor{teal}{$|$result 89.3333333333}
\textcolor{blue}{$|$output 89.33}

\end{mdframed}

\vspace{-5pt}
\caption{Example from MathQA, as a standard MathQA task and the corresponding tool-augmented sequence..}
\label{fig:mathqa_example}
\end{figure}

\begin{figure}[tbh]
\centering
\includegraphics[trim=0 0 50 50, width=0.9\linewidth]{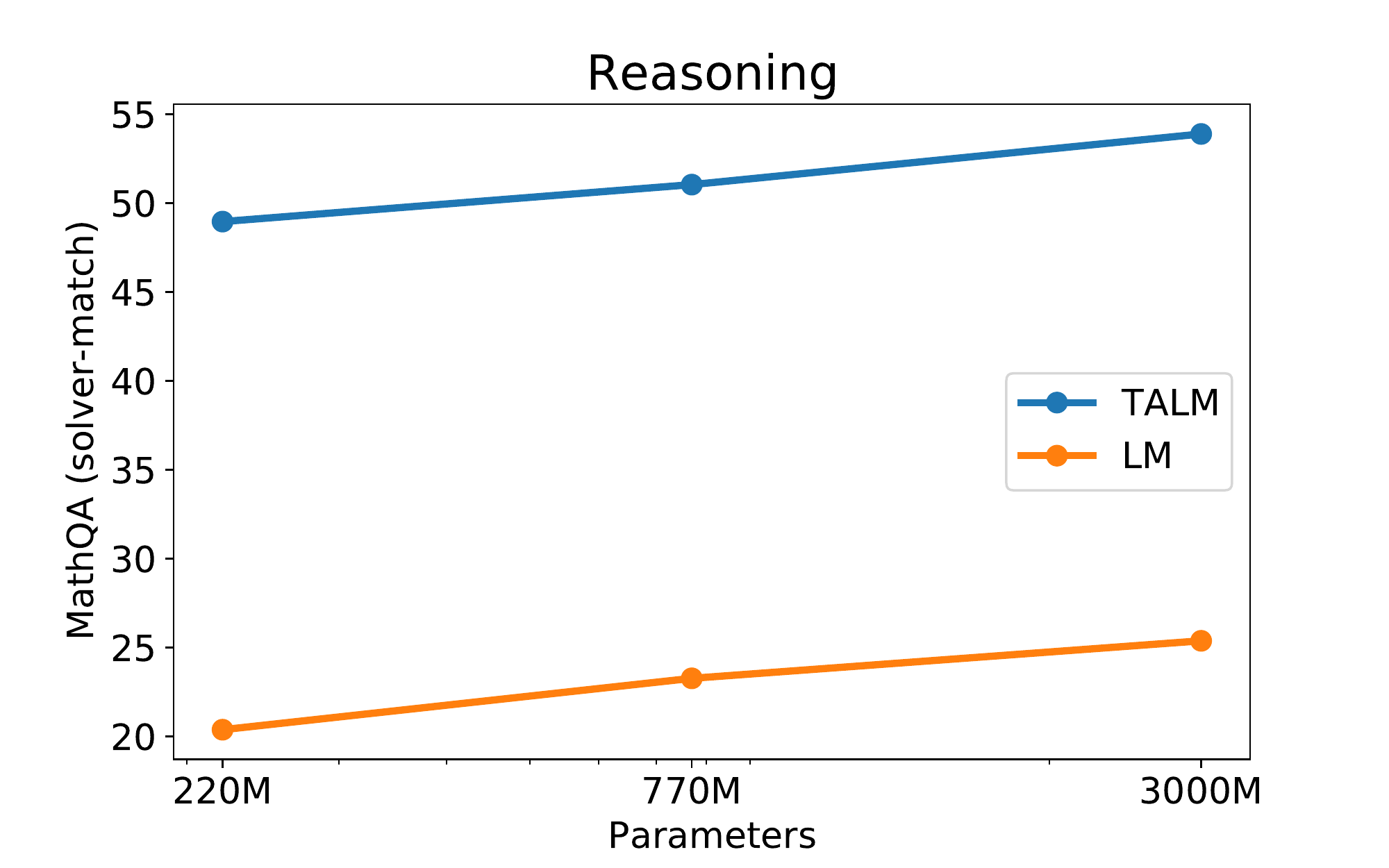}
\caption{Performance of TALM compared with LM of different model sizes on MathQA.}
\label{fig:mathqa_talm}
\end{figure}

We implemented a simple solver tool to execute formulas and check their results' correctness against their associated text-form answers. According to our solver tool, approximately $70\%$ of the formulas in MathQA produce results that match their corresponding answers, similar to the findings in \citep{DBLP:journals/corr/abs-2103-03874}. Our manual inspections show that mismatched results are due to either wrong formulas or invalid answers. The bootstrap tool-use dataset consists of a random sample of 10\% of the training corpus where the formula is valid ($\approx 2k$ examples). The TALM significantly outperforms a non-augmented LM as shown in Fig.~\ref{fig:mathqa_talm}.

\subsection{Self-Play Ablations}
\label{sec-self-play-ablations}

We find that TALMs perform significantly better after a single round of self-play than after training only on the limited bootstrap tool-use training examples, as shown in Fig.~\ref{fig:self-play-round}. Their performance continues to increase over three rounds of self-play. This trend holds across model sizes ranging from $220M$ to $3B$.

\subsection{Out-of-distribution Examples}

One benefit of TALM is its capability to generalize to input text that is out-of-distribution to the model's training data, yet solvable with access to tools.

On the knowledge-heavy QA task, we replace the BM-25 Wiki retriever with a public search engine, and show that TALM handles changing world knowledge well (see Fig.~\ref{fig:ood_search}).
\begin{figure}[H]
\centering

\begin{mdframed}
\small

\textbf{Question:} \textcolor{teal}{What is wordle?} 


\vspace{3pt}

\textbf{LM:}
\textcolor{violet}{a word generator} 
\vspace{3pt}

\textbf{TALM:}
\textcolor{violet}{a simple online word game that challenges people to find a five-letter word in six guesses}

\end{mdframed}
\vspace{-5pt}

\caption{LM vs TALM on changing knowledge.}
\label{fig:ood_search}
\end{figure}

On the math task, we test large number handling, an area where training data is lacking and non-augmented LMs are known to perform poorly \citep{brown2020gpt3}. See Fig.~\ref{fig:ood_math} demonstrating that TALM can handle large numbers, where a LM does not.
\begin{figure}[H]
\centering

\begin{mdframed}
\small




\textbf{Question:}
\textcolor{teal}{A car is driving 535 miles per hour, how many hours does it take to travel 2450 miles?} 
\vspace{3pt}

\textbf{LM:}
\textcolor{violet}{8.5} 
\vspace{3pt}

\textbf{TALM:}
\textcolor{violet}{4.58} 


\end{mdframed}
\vspace{-5pt}

\caption{LM vs TALM on a large number operation.}
\label{fig:ood_math}
\end{figure}

\section{Conclusion}
In this paper we present TALM, a framework for augmenting language models with arbitrary tools. TALM has two key ideas. First, we model tool-use via a text-to-text interface. Second, we apply an iterative self-play technique to bootstrap high performance on tasks with few tool-use labelled examples. Taken together, this interface and technique make exploring additional tools and tasks possible, without requiring expensive data labeling efforts.



TALM consistently outperforms a non-augmented LM on both a knowledge task (NQ) and reasoning task (MathQA). Ablations show that self-play is key to good performance, and that iterative self-play yields further gains. We conclude that the combination of tool augmentation and iterative self-play enables smaller models to outperform larger non-augmented LMs. 

We hope that this work enables further research into tool augmented language models, a promising direction to enhance model capabilities with less dependency on scale than many contemporary approaches.

\bibliography{references}

\bibliographystyle{plainnat}

\section{Appendix}

\subsection{Acknowledgements}
The authors would like to thank Noam Shazeer for early brainstorming on the path towards this work. We also thank Igor Mordatch for discussions and feedback. Finally we thank Mohammad Saleh for his helpful review and feedback improving this manuscript.

\subsection{Author Contributions}
This section lists the author contributions of each author.

\begin{itemize}
  
  \item Aaron Parisi designed and implemented tool-augmentation and self-play pipelines. Aaron ran the vast majority of experiments, and participated in brainstorming and paper writing.
  
  \item Yao Zhao participated in brainstorming, experimental setup discussion and paper writing. Yao implemented NQ/mathQA baselines and mathQA solvers.
  
  \item Noah Fiedel conceived of the project, participated in brainstorming, led the research group and writing the paper.
\end{itemize}


\end{document}